\documentclass[letterpaper, 10 pt, journal, twoside]{IEEEtran}
\ifCLASSINFOpdf
\else
\fi
\usepackage{times} 
\usepackage{amssymb}  
\usepackage{amsmath}
\usepackage{color}
\newtheorem{proposition}{Proposition}
\newtheorem{remark}{Remark}
\usepackage{graphicx}

\begin{document}
%
\title{A data-efficient geometrically inspired\\ polynomial kernel for robot inverse dynamics}

\author{Alberto Dalla Libera, and Ruggero Carli%
	\thanks{Manuscript received: May, 25, 2019; Revised August, 20, 2019; Accepted September, 15, 2019.}
	\thanks{This paper
		was recommended for publication by Editor Dezhen Song upon evaluation of the Associate Editor and Reviewers’ comments.}
	\thanks{Alberto Dalla Libera and Ruggero Carli are with the Deptartment of Information Engineering, University of Padova, Via Gradenigo 6/B, 35131 Padova, Italy
		{\tt\small dallaliber@dei.unipd.it}, {\tt\small carlirug@dei.unipd.it}}%
	\thanks{Digital Object Identifier (DOI): see top of this page.}}

%
%

\markboth{IEEE Robotics and Automation Letters. Preprint Version. Accepted September, 2019}
{Dalla Libera \MakeLowercase{\textit{et al.}}: Geometricaly Inspired Polynomila Kernel} 

%



\maketitle

\begin{abstract}
In this paper, we introduce a novel data-driven inverse dynamics estimator based on Gaussian Process Regression. Driven by the fact that the inverse dynamics can be described as a polynomial function on a suitable input space, we propose the use of a novel kernel, called \emph{Geometrically Inspired Polynomial Kernel} (GIP). 
The resulting estimator behaves similarly to model-based approaches as concerns data efficiency. Indeed, we proved that the GIP kernel defines a finite-dimensional Reproducing Kernel Hilbert Space that contains the inverse dynamics function computed through the Rigid Body Dynamics. The proposed kernel is based on the recently introduced \emph{Multiplicative Polynomial Kernel}, a redefinition of the classical polynomial kernel equipped with a set of parameters that allows for a higher regularization. We tested the proposed approach in a simulated environment, and also in real experiments with a UR10 robot. The obtained results confirm that, compared to other data-driven estimators, the proposed approach is more data-efficient and exhibits better generalization properties. Instead, with respect to model-based estimators, our approach requires less prior information and is not affected by model bias.
\end{abstract}


%
\IEEEpeerreviewmaketitle

\section{INTRODUCTION}  
\IEEEPARstart{L}{earning} the inverse dynamics model of a robot directly from data is still a challenging task in robotics, worth of investigation, as demonstrated by several important applications. For instance, by learning such a model, it is possible to design robot controllers based on feed-forward strategies \cite{feedforward} and on more complex Model Predictive Control approaches \cite{MPC_control}, or to provide  robots with proprioceptive sensing capabilities \cite{Collision_detection}, \cite{Force_control}.

%

Learning models directly from data has several advantages. Firstly, the derivation of a model is not always an easy task, and, even when a model is available, its use introduces a bias, due to uncertainties on the values of parameters which are assumed known, or to assumptions which are just a rough approximation of the real behavior of the robot.  
Secondly, data-driven approaches are not platform-dependent, namely, the same learning technique can be applied to different physical platforms, leading to considerable advantages in terms of design time and costs.

Several data-driven strategies to learn inverse dynamics have been developed. In \cite{SARCOS}, the authors proposed a locally weighted projection of different linear models. A significant number of approaches rely on neural networks;  for instance, the authors in \cite{RNN} resort to the use of a  recurrent neural network, while in \cite{LSTM} an LSTM network has been proposed. Another wide class of solutions is based on Gaussian Process Regression (GPR), \cite{GP_for_inv_dyn}, \cite{romeres}, \cite{Camoirano} and \cite{local_GPR}. Differently, from neural networks, GPR provides also a bound on the uncertainty of the estimates; this additional information can be exploited in different ways, see for instance in Reinforcement Learning the PILCO algorithm \cite{PILCO}.

Although data-driven modeling techniques have been applied successfully in several control applications, see for example \cite{PILCO}, \cite{GPS}, \cite{MPC_GP}, they are still not able to guarantee the same generalization properties of model-based learning techniques. Indeed, data-driven approaches capture only similarity between data, without exploiting important features, like causality or the presence of constraints imposed by physics and geometry. This fact results in a considerable data inefficiency, which is particularly evident in systems with a high number of degrees of freedom (DOF). The typical huge amount of data required by standard data-driven approaches poses serious limitations on their applicability, mainly due to the high computational burden needed to process all the available information, in addition to the difficulty of guaranteeing good generalization properties.

In this paper, we investigate the possibility of developing data-driven estimators of robot inverse dynamics exhibiting good generalization properties and high data efficiency. The main contribution of the paper is the design of a data-driven inverse dynamics estimator based on GPR, more precisely on a novel kernel function, named \emph{Geometrically Inspired Polynomial Kernel} (GIP). 
The main idea supporting our approach is related to the existence of a suitable transformation of the standard inputs, that are, positions, velocities, and accelerations, of the generalized coordinates, into an augmented space, where the inverse dynamics map derived with the Lagrangian equations is a polynomial function. Inspired by this property, we propose a model based on the \emph{Multiplicative Polynomial Kernel} (MPK), recently introduced in \cite{MPK}, which is a re-parameterization of the standard polynomial kernel. As shown in \cite{MPK}, compared to the standard polynomial kernel, the MPK parametrization allows for greater flexibility in neglecting eventual unnecessary basis functions of the corresponding Reproducing Kernel Hilbert Space (RKHS), leading to higher generalization performance.

The idea of mapping the standard inputs in a space where the inverse dynamics is a polynomial function has been explored also in \cite{Struct_net}. In \cite{Struct_net}, the authors have introduced the same augmented space we consider in this paper but have modeled only the elements of the inertia matrix and the potential energy as polynomials in this augmented space. However, instead of exploiting the encoding properties of polynomial kernels, they have designed a structural network inspired by the Lagrangian equations, where each element of the inertia matrix and the potential energy are a linear combination of monomials belonging to a particular class of polynomials. As consequence, compared to our approach, in the model proposed in \cite{Struct_net} the number of parameters to be identified grows quickly w.r.t. the robot DOF, i.e., w.r.t. the number of possible monomials, increasing the risk of overfitting.    

The polynomial-based strategy we introduce is tested both in a simulated environment and with data acquired from real experiments on a UR10 robot. Despite the GIP estimator requires minimal prior information compared to model-based estimators, the obtained results show that the proposed approach exhibits comparable performance in terms of accuracy and generalization. Additionally, compared to data-driven approaches, our learning  algorithm is more data-efficient and exhibits better generalization properties. 

The paper is organized as follows. In Section \ref{eq:learning_strategies}, we provide an overview of the main strategies based on GPR adopted in inverse dynamics learning. In Section \ref{sec:proposed_approach} we describe the approach we propose. Firstly, we identify an input transformation that leads to a description of the rigid body dynamics equations in terms of polynomial functions. Secondly, we briefly review MPK. Thirdly, we define the GIP kernel. Finally, in Section \ref{sec:experimental_results}, we test the proposed estimator in a simulated environment, representing a SCARA robot, and on data coming from real experiments performed with a UR10 robot.

\section{ROBOT INVERSE DYNAMICS: LEARNING STRATEGIES}\label{eq:learning_strategies}
In this section, we briefly review the dynamics model of robot manipulators and the main approaches proposed to deal with the inverse dynamics problem.

Consider a robot manipulator with $n+1$ links and $n$ joints, and let $\boldsymbol{q} =\left[q_1, \ldots, q_n\right]^T \in \mathbb{R}^n$ be the vector collecting the generalized coordinates associated to the joints; accordingly, let $\boldsymbol{\dot{q}}$ and $\boldsymbol{\ddot{q}}$ be the velocity and acceleration vectors, respectively.
The inverse dynamics problem consists in estimating the function mapping the triple $\left(\boldsymbol{q}, \boldsymbol{\dot{q}}, \boldsymbol{\ddot{q}}\right)$ into the vector of generalized torques, denoted by $\boldsymbol{\tau} \in \mathbb{R}^n$. The estimation is typically performed starting from a set of input-output observations, which is composed by the set of input locations $X = \left\{\boldsymbol{x}(t_1),\dots,\boldsymbol{x}(t_{N_{TR}})\right\}$, where $\boldsymbol{x}(t) = \left[\boldsymbol{q}(t) \,, \boldsymbol{\dot{q}}(t) \,,\boldsymbol{\ddot{q}}(t)\right]$, and the set of outputs $Y$, corresponding to the the noisy measures of $\left\{ \boldsymbol{\tau}(t_1), \dots, \boldsymbol{\tau}(t_{N_{TR}})\right\}$; $N_{TR}$ is the total number of observations. In the following, when there is no risk of confusion, we will omit the dependence on time $t$. 

\subsection{Rigid body dynamics estimators}

Several approaches which have been proposed to deal with the inverse dynamics problem are based on the rigid body dynamics (RBD) assumption. Under this assumption, the robot dynamics is described as
\begin{equation}
\boldsymbol{\tau} = B \left( \boldsymbol{q} \right) \boldsymbol{\ddot{q}} + C \left( \boldsymbol{q}, \boldsymbol{\dot{q}} \right) \boldsymbol{\dot{q}} + \boldsymbol{g} \left( \boldsymbol{q} \right) \text{,}
\label{eq:dyn}
\end{equation} 
where $B \left( \boldsymbol{q} \right) \in \mathbb{R}^{n \times n}$ and $C \left( \boldsymbol{q}, \boldsymbol{\dot{q}} \right) \in \mathbb{R}^{n \times n}$ are, respectively, the inertia matrix and the Coriolis matrix, and $\boldsymbol{g}\left(\boldsymbol{q}\right)$ is the vector accounting for the gravitational contributions, see \cite{Siciliano}. The previous equation depends on two sets of parameters, the kinematic and dynamics parameters. The first set is composed of the geometric quantities (i.e., lengths, angles) that, together with $\boldsymbol{q}$, define the forward kinematic.
The second set, instead, contains the masses, centers of mass, and inertia components of the links. Remarkably, it is possible to show that \eqref{eq:dyn} is linear w.r.t. the dynamics parameters, see \cite{Siciliano}. Specifically, denoting by $\boldsymbol{w}$ the vector collecting all the dynamics parameters, 
\eqref{eq:dyn} can be rewritten as
\begin{eqnarray}
\boldsymbol{\tau} &=& \Phi\left(\boldsymbol{q}, \boldsymbol{\dot{q}}, \boldsymbol{\ddot{q}} \right) \boldsymbol{w} = 
\Phi\left( \boldsymbol{x} \right) \boldsymbol{w}\nonumber\\
&=&\begin{bmatrix}
\boldsymbol{\phi}_1^T\left(\boldsymbol{x} \right) && \dots && \boldsymbol{\phi}_n^T\left(\boldsymbol{x} \right)
\end{bmatrix}^T \boldsymbol{w}
\text{,} \label{eq:dyn_lin}
\end{eqnarray}
for a suitable matrix $\Phi \in \mathbb{R}^{n \times N_{par}}$, which depends only on the kinematic parameters. Then, assuming the kinematic parameters to be known, the inverse dynamics problem boils down to the computation of $\hat{\boldsymbol{w}}$, an estimate  of $\boldsymbol{w}$. 

In several solutions, $\hat{\boldsymbol{w}}$ is computed relying on Fisherian techniques, see for example \cite{Sousa}. When the model is accurate, and the signal to noise ratio is sufficiently high, these estimators achieve accurate estimates, together with good generalization properties. However, besides the presence of noise, several aspects could limit the performance of such approaches. Indeed, it is worth stressing that errors about the kinematic parameters introduce model bias. Moreover, there are situations where it is hard to derive \eqref{eq:dyn_lin}, or where the rigid body assumption is a too rough approximation of the real robot behaviors.


\subsection{Gaussian Process Regression for robot inverse dynamics}
To overcome the limitations characterizing estimators based on the RBD assumption, several Bayesian approaches have been proposed in the last decade.
Most techniques are based on GPR, see \cite{Rasmussen} for a detailed description. Typically, in GPR approaches, each joint is treated individually and modeled as a single Gaussian Process. More precisely, when considering the $\emph{i-th}$ joint, it is assumed that the output measurements $\boldsymbol{y}_i= \left\{\tau_i\left(t_1\right)\,,\dots\,, \tau_i\left(t_{N_{TR}}\right) \right\}$ are generated by the following probabilistic model
\begin{equation}
\boldsymbol{y}_i = \begin{bmatrix}
f_i\left(\boldsymbol{x}\left(t_1\right)\right) \\ \vdots \\ f_i\left(\boldsymbol{x}\left(t_{N_{TR}}\right)\right)
\end{bmatrix} +
\begin{bmatrix}
e_i\left(t_1\right) \\ \vdots \\ e_i\left(t_{N_{TR}}\right)
\end{bmatrix}  
= \boldsymbol{f}_i\left(X\right) + \boldsymbol{e}_i
\label{eq:GPR_model} \text{,}
\end{equation}
where $\boldsymbol{e}_i$ is i.i.d. Gaussian noise with standard deviation $\sigma_{n_i}$, and $\boldsymbol{f}_i\left(X\right)$ is an unknown function defined as a Gaussian Process, namely, $\boldsymbol{f}_i\left(X\right) \sim N\left(\boldsymbol{m}_i\left(X\right)\, , K_i\left(X,X\right)\right)$, being $\boldsymbol{m}_i$ and $K_i\left(X,X\right)$, respectively, mean and covariance. In particular, the matrix $K_i\left(X,X\right)$, called also \emph{kernel matrix}, is defined through a kernel function $k_i\left(\cdot, \cdot\right)$, i.e., the element in $h$-th row and $j$-th column is equal to $k_i\left(\boldsymbol{x}\left(t_h\right), \boldsymbol{x}\left(t_j\right)\right)$.

In this probabilistic framework, the maximum a posteriori probability (MAP) estimator of $f_i$ is given in closed form. Let $\boldsymbol{x}_{*}$ be a general input location. Then, as proved in \cite{Rasmussen} (see Chapter 2), the MAP estimate of $f_i(\boldsymbol{x}_{*})$ is given by
	\begin{equation}\label{eq:meanGP}
	\hat{f}_i(\boldsymbol{x}_*) = \sum_{h=1}^{N_{TR}} \alpha_i k_i(\boldsymbol{x}_*, \boldsymbol{x}(t_h)) \text{,}
	\end{equation}
	where $\alpha_i$ is the \emph{i}-th element of the vector $\boldsymbol{\alpha}$, defined as 
	\begin{equation}\label{eq:alphaGP}
	\boldsymbol{\alpha} = \left(K(X,X)+\sigma_{n_i}^2 I\right)^{-1} \boldsymbol{y}_i \text{.}
	\end{equation}
	Before providing a brief overview about GPR based solutions, we comment about the computational cost of \eqref{eq:meanGP} and \eqref{eq:alphaGP}. Due to the matrix inversion in \eqref{eq:alphaGP}, the number of operations required to derive the $f_i$ estimator is proportional to the cube of $N_{TR}$. However, observe that $\boldsymbol{\alpha}$ depends only on the training data, then its computation can be performed offline. As consequence, once $\boldsymbol{\alpha}$ is computed, the cost of evaluating \eqref{eq:meanGP}, grows linearly w.r.t. $N_{TR}$, thus allowing the use of GPR estimators in real time applications, see for example \cite{MPC_GP} and \cite{local_GPR}.


When no prior knowledge about the model is available, the GPR prior can be defined in a data-driven way. An option consists in assuming that the distribution of the outputs is stationary with respect to the locations of the inputs, and then, considering $\boldsymbol{m}_i\left(\cdot\right)=0$, and adopting a Radial Basis Function (RBF) as kernel. The RBF kernel is  defined as 
	\begin{equation*}
	k_{RBF}(\boldsymbol{x}\left(t_h\right), \boldsymbol{x}\left(t_j\right)) = \lambda_{RBF} e^{||\boldsymbol{x}\left(t_h\right)-\boldsymbol{x}\left(t_j\right)||^2_{\Sigma^{-1}_{RBF}}} \text{,}
	\end{equation*}
	where $\lambda_{RBF}$ and $\Sigma_{RBF}$ are the  kernel hyperparameters, which are typically tuned from data by Marginal Likelihood (ML) maximization.  
In general, estimators based on RBF kernels well approximate the inverse dynamics only in a neighborhood of the training input locations. Several strategies have been designed to limit the computational complexity and to increase the generalization, see \cite{local_GPR}. Despite that, when considering robots with a considerable number of DOF, it is still hard to design inverse dynamics estimators based on RBF kernel with remarkable generalization properties.


In case a RBD model is given, starting from \eqref{eq:dyn_lin}, and modeling $\boldsymbol{w}$ as a Gaussian variable, with mean $\bar{\bf{w}}$ and covariance $\Sigma_w$, it is possible to derive a linear kernel that inherits all the positive aspects of the RBD estimators, but acts in a Bayesian framework. Consider joint \emph{i}, then
\begin{eqnarray}
&\boldsymbol{m}_i (X) = \Phi_i (X)\bar{\bf{w}} \nonumber\\
&k_{i}(\boldsymbol{x}\left(t_h\right), \boldsymbol{x}\left(t_j\right)) = \boldsymbol{\phi}_i\left(\boldsymbol{x}\left(t_h\right)\right) \Sigma_w \boldsymbol{\phi}_i^T\left(\boldsymbol{x}\left(t_j\right)\right) \label{eq:PP_kernel} \text{,}
\end{eqnarray}
where $\Phi_i (X)$ is the matrix collecting all the rows $\boldsymbol{\phi}_i\left(\boldsymbol{x}\left(t_j\right)\right)$, $j=1, \ldots, N_{TR}$, and where, in this case, $\bar{\bf{w}}$ and $\Sigma_w$ are the kernel hyperparameters. The above kernel can be used alone, leading to the so-called Parametric Prior (PP) estimators, or together with a data-driven kernel, leading to the so-called Semiparametric Prior (SP) estimators. In the latter case, when adopting the RBF kernel as data-driven kernel, we have
\begin{eqnarray}
&&k_i(\boldsymbol{x}\left(t_h\right), \boldsymbol{x}\left(t_j\right)) =  \label{eq:SP_kernel}\\
&&\qquad \boldsymbol{\phi}_i\left(\boldsymbol{x}\left(t_h\right)\right) \Sigma_w \boldsymbol{\phi}_i^T\left(\boldsymbol{x}\left(t_j\right)\right)+ k_{RBF}(\boldsymbol{x}\left(t_h\right), \boldsymbol{x}\left(t_j\right)), \nonumber
\end{eqnarray}
see for example \cite{romeres}, \cite{Camoirano} and \cite{Peters}. The rationale behind the use of kernel in \eqref{eq:SP_kernel} is the following: the first term allows exploiting the prior knowledge coming from the RBD, providing generalization, while $k_{RBF}\left(\cdot,\cdot\right)$ improves the estimate in a neighborhood of the training locations, compensating for model bias or un-modeled behaviors.
We remark that estimators based on \eqref{eq:PP_kernel} and \eqref{eq:SP_kernel} are model-based estimators, since their kernel functions are derived starting from \eqref{eq:dyn_lin}.

\section{PROPOSED APPROACH: GEOMETRICALLY INSPIRED POLYNOMIAL KERNEL} \label{sec:proposed_approach}
In this section, we propose a novel kernel that allows estimating the inverse dynamics without requiring prior knowledge of the model, preserving good generalization and high accuracy. This section is organized as follows. Firstly, we state Proposition \ref{prop:inv_dyn_poly}, which characterizes the inverse dynamics from the functional analysis point of view. Given the type of each joint, i.e. prismatic or revolute, Proposition \ref{prop:inv_dyn_poly} defines a transformation of the input $\boldsymbol{x}$ where the inverse dynamics is a polynomial function. Then, we briefly review the \emph{Multiplicative Polinomial Kernel}, and, finally, we define the proposed kernel function, named \emph{Geometrically Inspired Polynomial kernel}.

\subsection{Polynomial characterization of the rigid-body model}

In the following, we restrict our study to manipulators where each joint is either revolute or prismatic. 
Let $N_r$ and $N_p$ be the number of revolute and prismatic joints, respectively, where $N_r+N_p=n$, and let us denote by $I_r = \left\{ i_{r_1} , \dots , i_{r_{N_r}} \right\}$ and $I_p = \left\{ i_{p_1} , \dots , i_{p_{N_p}} \right\}$ the sets containing, respectively, the revolute and prismatic joints indexes.
We start our analysis by defining
\begin{eqnarray*}
	&&\boldsymbol{q}_{c} = 
	\left[ \cos\left(q_{i_{r_1}}\right) \,,\, \dots \,,\, \cos\left(q_{i_{r_{N_r}}}\right) \right] \in \mathbb{R}^{N_r}
	\text{,} \\
	&&\boldsymbol{q}_{s} = 
	\left[\sin\left(q_{i_{r_1}}\right) \,,\, \dots \,,\, \sin\left(q_{i_{r_{N_r}}}\right) \right] \in \mathbb{R}^{N_r} \text{,}\\
	&&\boldsymbol{q}_{p} = 
	\left[ q_{i_{p_1}} \,,\, \dots \,,\, q_{i_{p_{N_p}}} \right] \in \mathbb{R}^{N_p} \text{.}
\end{eqnarray*}
In the following, we denote by $q_{c_b}$ the element in $\boldsymbol{q}_{c}$ associated to joint $i_{r_b}$, i.e. $\cos(q_{i_{r_b}})$ (similar definitions hold for $q_{s_b}$ and $q_{p_b}$). For later convenience we define also $\boldsymbol{q}_{cs} \in \mathbb{R}^{2N_r}$, the vector obtained concatenating $\boldsymbol{q}_{c}$ and $\boldsymbol{q}_{s}$.
In addition, we denote by $\boldsymbol{\dot{q}}_{v}$ the vector collecting the elements of the set
$$
\left\{ \dot{q}_i\dot{q}_j ,\, 1\leq i \leq n \,,\,i\leq j \leq n \right\},
$$
that is the set containing all the possible pairwise products of components of $\boldsymbol{\dot{q}}$. Notice that $\boldsymbol{\dot{q}}_{v} \in \mathbb{R}^{n\left(n+1\right)/2}$.

Finally, we introduce a compact notation to identify a particular set of inhomogeneous polynomial functions. Let $\boldsymbol{a}$ be the vector containing the $m$ variables $a_1,\ldots, a_m$. We denote by $\mathbb{P}_{[p]}\left(\boldsymbol{a}_{[p_1]}\right)$ the set of polynomial functions of degree not greater than $p$ defined over the variables in $\boldsymbol{a}$, such that each variable $a_i$ appears with degree not greater than $p_1$. Similar definitions hold in case the inputs set accounts for more input vectors. 

Now we consider the transformation $F : \mathbb{R}^{3n} \to \mathbb{R}^\gamma$, with $\gamma= 2N_r + N_p + n(n+1)/2 + n$, which maps the input location $\boldsymbol{x}$ into the element $\bar{\boldsymbol{x}} \in \mathbb{R}^\gamma$, defined as
\begin{equation}
\bar{\boldsymbol{x}} = \left[\boldsymbol{q}_{c}, \boldsymbol{q}_{s}, \boldsymbol{q}_{p}, \boldsymbol{\dot{q}}_{v}, \boldsymbol{\ddot{q}}\right] \text{.} \label{eq:input_space}
\end{equation}
We have the following result.


\begin{proposition} \label{prop:inv_dyn_poly}
	Consider a manipulator with $n+1$ links and $n$ joints, divided in $N_r$ revolute joints and $N_p$ prismatic joints, subject to $n=N_r+N_p$. Then, the inverse dynamics of each joint obtained through the rigid body model in \eqref{eq:dyn} belongs to $\mathbb{P}_{(2n+1)}\left(\boldsymbol{q}_{c(2)}, \boldsymbol{q}_{s(2)}, \boldsymbol{q}_{p(2)}, \boldsymbol{\dot{q}}_{v(1)}, \boldsymbol{\ddot{q}}_{(1)}\right)$. Namely, each $\tau_i\left(\cdot\right)$ is a polynomial function in $\bar{\boldsymbol{x}}$, of degree not greater than $2n+1$, such that: (i) each element of $\boldsymbol{q}_c$, $\boldsymbol{q}_s$ and $\boldsymbol{q}_p$ appears with degree not greater than $2$, and (ii) each element of $\boldsymbol{\dot{q}}_v$ and $\boldsymbol{\ddot{q}}$ appears with degree not greater than $1$. Moreover, for any monomial of the aforementioned polynomial, the sum of the $q_{c_b}$ and $q_{s_b}$ degrees is equal or lower than two, namely, it holds
	\begin{equation*}
	deg\left(q_{c_b}\right)+deg\left(q_{s_b}\right) \leq 2\text{ .}
	\end{equation*}
\end{proposition}

\vspace{0.3cm}
The proof is reported in the Appendix.
\begin{remark}
	The result stated in Proposition \ref{prop:inv_dyn_poly} is related to the modeling property used in \cite{Struct_net}, though some important differences are present. Indeed, in  \cite{Struct_net}, the authors have modeled with polynomial functions the potential energy and the elements of the inertia matrix, while Proposition \ref{prop:inv_dyn_poly} establishes that the whole inverse dynamics is a polynomial function in the augmented space. Moreover, Proposition \ref{prop:inv_dyn_poly} provides more strict constraints on the degrees of $\cos(q_i)$,
		$\sin(q_i)$, thus better characterizing  the maximum degree with which each variable can appear in the different monomials, and, in turn, decreasing the number of possible monomials.
\end{remark}

\subsection{Multiplicative Polynomial Kernel}
From a functional analysis point of view, Proposition \ref{prop:inv_dyn_poly} states that the inverse dynamics derived through the RBD belongs to the finite dimensional space of polynomial functions. A suitable set of basis functions for this space is given by the set of all the monomials in $\mathbb{P}_{(2n+1)}\left(\boldsymbol{q}_{c(2)}, \boldsymbol{q}_{s(2)}, \boldsymbol{q}_{p(2)}, \boldsymbol{\dot{q}}_{v(1)}, \boldsymbol{\ddot{q}}_{(1)}\right)$, heareafter denoted by vector $\boldsymbol{\phi}_{\mathbb{P}}\left(\boldsymbol{x}\right) \in \mathbb{R}^{N_{\mathbb{P}}}$, being $N_{\mathbb{P}}$ its cardinality. Unfortunately, $N_{\mathbb{P}}$ grows rapidly with the number of joints. To provide a couple of examples, when considering a SCARA robot, i.e. $n=4$, we have that $N_{\mathbb{P}} = 1647$, while, for a standard six DOF robot, like the UR10, we have that $N_{\mathbb{P}} = 302615$. Clearly, when considering GPR based approaches, the computational and memory requirements induced by the dimension of $\boldsymbol{\phi}_{\mathbb{P}}\left(\boldsymbol{x}\right)$ prevent the possibility of adopting the linear kernel $k\left(\boldsymbol{x}\left(t_k\right),\boldsymbol{x}\left(t_j\right)\right)=\boldsymbol{\phi}_{\mathbb{P}}\left(\boldsymbol{x}\left(t_j\right)\right) \Sigma_w \boldsymbol{\phi}^T_{\mathbb{P}}\left(\boldsymbol{x}\left(t_j\right)\right)$, as done in \eqref{eq:PP_kernel}.


A compact solution that allows overcoming this problem consists in assuming that the target function $\tau_i\left(\cdot\right)$ belongs to the Reproducing Kernel Hilbert Space (RKHS) associated to a polynomial kernel, see \cite{Rasmussen}. More precisely, when considering the space of inhomogeneus polynomials defined on the components of $\boldsymbol{x} \in \mathbb{R}^{d}$, with maximum degree $p$, the polynomial kernel is classically defined as
\begin{equation}
k^{(p)}_{pk}\left(\boldsymbol{x}\left(t_h\right),\boldsymbol{x}\left(t_j\right)\right) = \left(\sigma_p^2 + \boldsymbol{x}^T\left(t_h\right) \Sigma_p \boldsymbol{x}\left(t_j\right)\right)^p \text{,} \label{eq:poly_kernel}
\end{equation}
where $\sigma^2_p >0$ and $\Sigma_p >0$ are the kernel hyperparameters, see \cite{Rasmussen}. Unfortunately, as highlighted in \cite{Rasmussen} (chapter 4.2.2), the kernel function in \eqref{eq:poly_kernel} is not widely used in regression problems, since it is prone to overfitting, in particular when considering high dimensional inputs and $p>2$, that is exactly the situation identified in Proposition \ref{prop:inv_dyn_poly}. 

A valid alternative to \eqref{eq:poly_kernel} is represented by the MPK, recently introduced in \cite{MPK}. When considering the space of inhomogeneous polynomial with maximum degree $p$, the MPK is defined as the product of $p$ linear kernels,
\begin{equation}
k^{(p)}_{mpk}\left(\boldsymbol{x}\left(t_h\right),\boldsymbol{x}\left(t_j\right)\right) = \prod_{s=1}^{p}\left(\sigma^2_s+\boldsymbol{x}^T\left(t_h\right) \Sigma_{s} \boldsymbol{x}\left(t_j\right) \right) \text{,} \label{eq:poly_kernel_prop}
\end{equation}
where the $\Sigma_{s}>0 \in \mathbb{R}^{d\times d}$ matrices are distinct diagonal matrices. The diagonal elements, together with the parameters $\sigma^2_{s}$, compose the hyperparameters set of the MPK.

Observe that the RKHSs identified by \eqref{eq:poly_kernel_prop} and \eqref{eq:poly_kernel} contains the same basis functions. However, as discussed in \cite{MPK},  \eqref{eq:poly_kernel_prop} is equipped with a richer set of hyperparameters, that can be tuned by ML maximization, and allows a better selection of the monomials that highly influence the system output. 

\begin{figure}
	\vspace{0.2cm}
	\centering
	\includegraphics[width=1\linewidth]{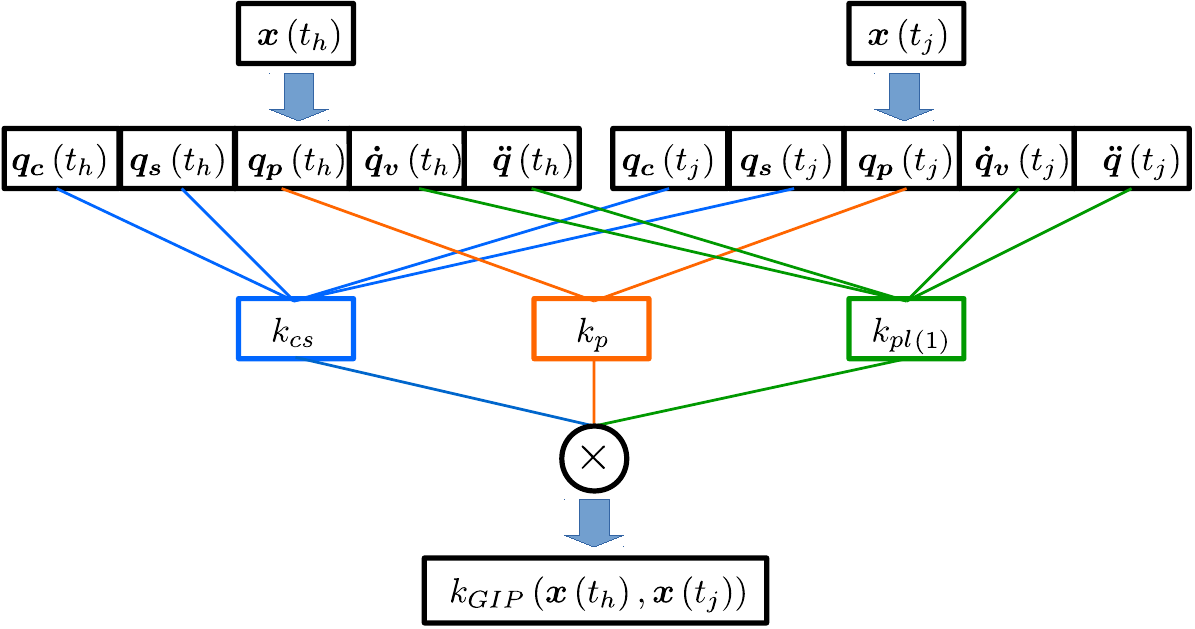}
	\caption{Schematic representation of the GIP kernel.}
	\label{fig:GIP_kernel}
\end{figure}

\subsection{Geometrically inspired polynomial kernel}
In this subsection, we describe the GIP kernel we propose to model the robot inverse dynamics. Our approach requires minimal information since we assume to know only the joints type. We model each joint torque with a zero-mean Gaussian process, and, driven by Proposition \ref{prop:inv_dyn_poly}, we assume that the inverse dynamics is a polynomial in the input space $\bar{\boldsymbol{x}}$, defined in \eqref{eq:input_space}. To comply with the constraints on the maximum degree of each term, we adopt a kernel function given by the product of $N_r+N_p+1$ kernels of the type defined in equation \eqref{eq:poly_kernel_prop}, where
\begin{itemize}
	\item $N_r$ kernels have $p=2$ and each of them is defined on a $2$-dimensional input space given by $\boldsymbol{q}_{cs_b} = \left[q_{c_b}, q_{s_b}\right]$, with $b \in I_r$;
	\item $N_p$ kernels have $p=2$ and each of them is defined on a $1$-dimensional input, given by one of the $\boldsymbol{q}_{p}$ components;
	\item  a single kernel with $p=1$ defined on the input vector $\boldsymbol{q}_{av} = \left[\boldsymbol{\ddot{q}}\,,\,\boldsymbol{\dot{q}}_{v}\right]$.  
\end{itemize}
The resulting kernel for the $i$-th joint is
\begin{eqnarray}
k_i\left(\bar{\boldsymbol{x}}\left(t_h\right),\bar{\boldsymbol{x}}\left(t_j\right)\right) &=& k_{cs} \left(\boldsymbol{q}_{cs}\left(t_h\right),\boldsymbol{q}_{cs}\left(t_j\right)\right) \notag \\
&\,&k_p \left(\boldsymbol{q}_{p}\left(t_h\right),\boldsymbol{q}_{p}\left(t_j\right)\right) \notag \\
&\,&k^{(1)}_{mpk}\left(\boldsymbol{q}_{av}\left(t_h\right),\boldsymbol{q}_{av}\left(t_j\right)\right) \text{,}\label{eq:GIP_kernel}
\end{eqnarray}
with
\begin{align*}
&k_{cs} \left(\boldsymbol{q}_{cs}\left(t_h\right),\boldsymbol{q}_{cs}\left(t_j\right)\right) = 
\prod_{b=1}^{N_r} k^{(2)}_{mpk}\left(\boldsymbol{q}_{cs_b}\left(t_h\right),\boldsymbol{q}_{cs_b}\left(t_j\right)\right) \text{,} \\
&k_p \left(\boldsymbol{q}_{p}\left(t_h\right),\boldsymbol{q}_{p}\left(t_j\right)\right) = 
\prod_{b=1}^{N_p} k^{(2)}_{mpk}\left(q_{p_b}\left(t_h\right),q_{p_b}\left(t_j\right)\right) \text{.}
\end{align*}
In Figure \ref{fig:GIP_kernel} we reported a schematic representation of the GIP kernel.

\section{EXPERIMENTAL RESULTS}\label{sec:experimental_results}
We tested the proposed approach both in a simulated environment and in a real environment. Regarding technical aspects, we implemented all the considered algorithms in Python. To speed up algebraic operations, we largely exploited the functionalities provided by Pytorch \cite{pytorch}. The code\footnote{https://bitbucket.org/AlbertoDallaLibera/gip\_kernel} and the datasets\footnote{https://mega.nz/\#F!fbBBSCCK!NwRs60ace05mTe2Ot5fz-Q} are publicly available.     

\subsection{Simulated SCARA robot}
To evaluate the benefits of the GIP kernel, we first tested the proposed approach in a simulated environment. We considered a SCARA robot, more precisely an AdeptOne Robot. The SCARA is a $4$ DOF robot manipulator, with three revolute joints (joint $1$, $2$ and $4$) and a prismatic joint (joint $3$). As far as data generation is concerned, joint torques were computed through \eqref{eq:dyn}, assuming complete knowledge of the joint trajectories. Equation \eqref{eq:dyn} was derived using the python package \textit{Sympybotics}\footnote{https://github.com/cdsousa/SymPyBotics}. 

\begin{figure}
	\centering
	\includegraphics[width=1\linewidth]{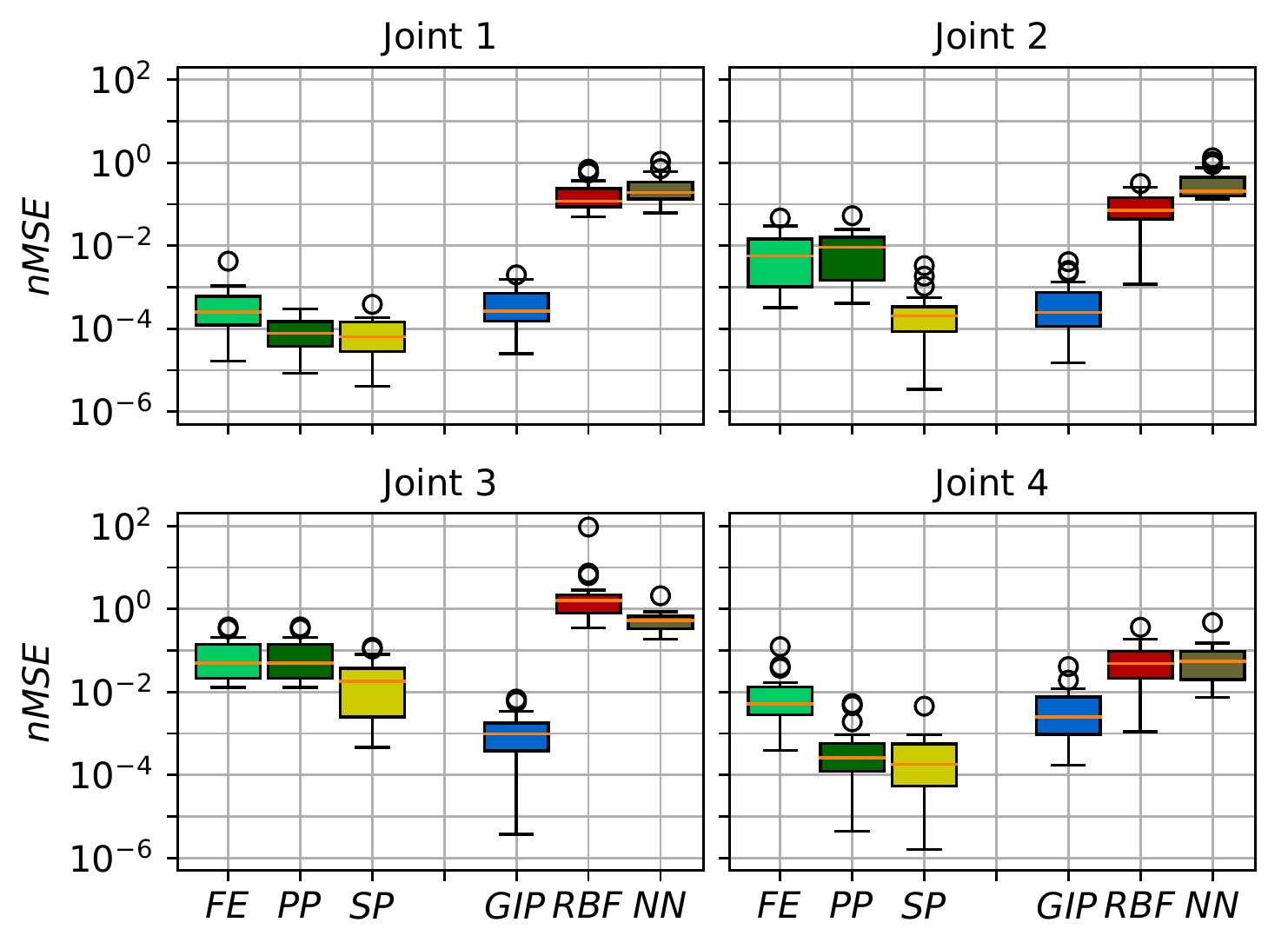}
	\caption{Boxplot of the $nMSE$s obtained in $20$ experiments. Model-based estimators (FE, PP, and SP) are on the left, while the data-driven estimators (GIP, RBF, and NN, the fully connected neural network) are on the right. The upper and the lower vertical edge of the box represent respectively the first and the third quartile, while the median is pointed out by the orange line. The black vertical lines indicate variability outside the first and third quartile; circles point out outliers. Results are plotted in logarithmic scale.}
	\label{fig:kin_error_robustness}
\end{figure}

\subsubsection{Estimation accuracy}
In the first experiment, we tested the estimators' accuracy. The proposed approach is compared with both model-based and data-driven estimators. As far as model-based estimators are concerned, we implemented three solutions. The first estimator is a classical Fisherian estimator (FE), based on \eqref{eq:dyn_lin}; in particular, we considered the implementation provided by \textit{Sympybotics}, see \cite{Sousa} for details. The other two model-based estimators are the PP and SP kernel-based estimators, where the model-based component is defined as in \eqref{eq:PP_kernel}. For all the three solutions, we computed the  $\Phi$ matrix in \eqref{eq:dyn_lin} assuming the nominal kinematic parameters provided by the manufacturer. To account for behaviors due to model bias, we varied the kinematic parameters of the model generating data around the nominal values, so that the $\Phi$ matrix used by the FE and the PP and SP kernels is different from the one generating data. 
Parameters perturbations are uniform random variables, ranging $\left[-0.05, 0.05\right] m$ for lengths, and $\left[-5, 5\right] deg$ for angles. Instead, as far as data-driven approaches are concerned, we tested an RBF kernel-based estimator and a neural network. The neural network is a fully connected network with two hidden layers, each of which is composed of $400$ sigmoids. Recurrent architectures have not been considered since typically they are applied to deal with the on-line adaptation problem, which is out of the scope of this work.

To obtain statistically significant results, we performed a Monte Carlo analysis, composed of $20$ experiments. A single experiment consists in sampling a model perturbation and performing two simulations, one for the training set and one for the test set. In each simulation, joints follow a distinct random trajectory, given by the sum of $200$ sinusoids with random amplitudes and angular velocities sampled in the range $[-2,2]\,rad/sec$. A zero-mean Gaussian noise with standard deviation $0.01 \,Nm$ was added to the output of \eqref{eq:dyn}, resulting in a high signal to noise ratio. Both the training and test dataset are composed of $2000$ samples.

The hyperparameters of the GPR based estimators were trained by ML maximization (see \cite{Rasmussen}, chapter 2.2, for a detailed discussion). As concerns $\boldsymbol{\bar{w}}$ and $\Sigma_w$, i.e., the prior distribution of the model-based kernels, we considered $\boldsymbol{\bar{w}}=0$, and  $\Sigma_w$ equals to a diagonal matrix with distinct diagonal elements. Instead, as concerns the optimization of neural network parameters, the Mean Squared Error (MSE) was considered as loss function, defined as
\begin{equation*}
MSE(\boldsymbol{y}_i,\boldsymbol{\hat{y}}_i) = \sum_{j=1}^{N}\left(y_i(t_j)-\hat{y}_i(t_j)\right)^2/N\text{,}
\end{equation*} 
with $N$ equals to the number of samples. Both for GPRs and the neural network, we used Adam as optimizer \cite{adam}. Further details about the optimization, size of the batch and number of epochs, are provided in the source code.

Performance are compared by Normalized Mean Squared Error ($nMSE$) in the test set, defined as
\begin{equation*}
nMSE(\boldsymbol{\tau}_i, \boldsymbol{\hat{y}}_i) = MSE(\boldsymbol{\tau}_i, \boldsymbol{\hat{y}}_i)/ Var\left(\boldsymbol{\tau}_i\right)\text{.}
\end{equation*} 

In Figure \ref{fig:kin_error_robustness}, we have plotted the obtained $nMSE$s through a boxplot. Results show that the proposed approach outperforms other data-driven estimators, which are not able to learn accurately the inverse dynamics of the SCARA robot using just $2000$ samples. Indeed, except for joint 4, the $nMSE$s of the RBF kernel-based estimator and the neural network estimator are in most of the trials higher than $10\%$. Instead, the GIP kernel-based estimator provides accurate estimates, as proven by $nMSE$s values, that are always below $1\%$, with the exception of joint 4, where two outliers are present, probably due to training inputs not sufficiently exciting. Moreover, the GIP kernel-based estimator performs similarly to the model-based approaches. Actually, in joint $2$ and $3$, the proposed approach outperforms FE and the PP kernel-based estimator, whose performances are affected by model bias. Results confirm also the validity of semiparametric schemes, proving that the addition of a data-driven component can compensate for model bias, given that the SP kernel-based estimator outperforms PP. Anyway, we highlight that in hybrid schemes the RBF component might not be effective in compensating for model bias, in particular when the performance of the data-driven estimator is low, as proven by the $nMSE$s in joint $3$, where the GIP kernel-based estimator is more accurate than SP. Finally, a comparison of the FE and PP performance suggests that Fisherian approaches are more sensitive to model bias; notice, in particular, the $nMSE$s obtained in joint 1 and 4.
\subsubsection{Data efficiency}
In the second experiment, we tested the data efficiency of different estimators. Since our focus is on comparing data-driven approaches, model bias was not considered, in favor of greater results interpretability. The GIP kernel-based estimator is compared with the other data-driven estimators, and also with the PP kernel-based approach. In this ideal scenario, where data are generated with the robot nominal parameters, the performance of PP might be considered as the baseline of an optimal solution.

\begin{figure}
	\vspace{0.2cm}
	\centering
	\includegraphics[width=1\linewidth]{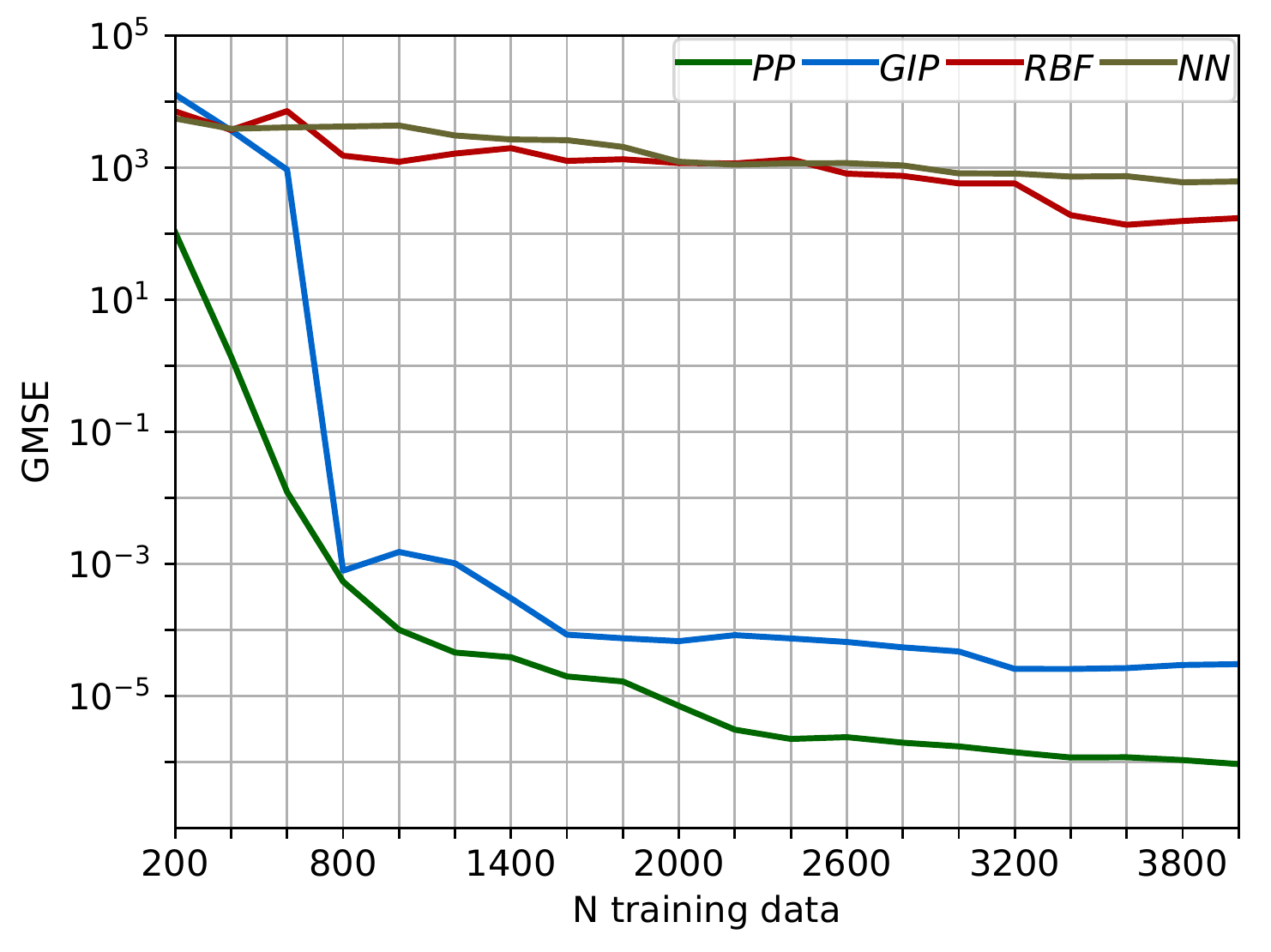}
	\caption{Comparison of the neural network (NN) and the PP, RBF and GIP kernel based estimators in terms of accuracy and data efficiency. The plot shows the evolution of $GMSE$ in the test set, as function of the training samples available. Results are plotted in logarithmic scale.}
	\label{fig:data_efficiency}
\end{figure}

The experiment is composed of training and test simulation, with joints trajectories generated as in the previous experiment; each dataset accounts for $4000$ samples. Results are reported in Figure \ref{fig:data_efficiency}, where we have plotted the evolution of the Global Mean Squared Error ($GMSE$), i.e., the sum of the $MSE(\boldsymbol{\tau}_i,\boldsymbol{\hat{y}}_i)$s of  the four joints, as function of the number of training samples used to train and derive estimators. 
The evolutions of the errors show that the proposed solution outperforms the other data-driven estimators, both in terms of accuracy and data efficiency, given that its $GMSE$ is lower and decreases faster. As in the previous experiment, the GIP kernel-based estimator behaves more similarly to the model-based approach than to the other data-driven solutions, proving its data efficiency. 

\subsection{UR10 robot}
We used a Universal Robots UR10 to test the proposed approach in a real setup. The UR10 robot is a $6$ DOF collaborative manipulator, where all the joints are revolute. This robot is not equipped with joint torque sensors, but one can directly measure the motor currents $\boldsymbol{i}$. Assuming that the behaviors due to elasticity are negligible, i.e. $\boldsymbol{\theta}=K_r\boldsymbol{q}$, where $\boldsymbol{\theta}$ contains the motor angles, and $K_r$ is the diagonal matrix of the gear reduction ratio, the inverse dynamics in \eqref{eq:dyn} can be rewritten as
\begin{equation*}
K_{eq} \boldsymbol{i} = B_{eq} \left( \boldsymbol{q} \right) \boldsymbol{\ddot{q}} + C \left( \boldsymbol{q}, \boldsymbol{\dot{q}} \right) \boldsymbol{\dot{q}} + \boldsymbol{g} \left( \boldsymbol{q} \right) + F_v \boldsymbol{\dot{q}} + F_c sign\left(\boldsymbol{\dot{q}}\right)\text{,}
\end{equation*}
where $F_v + F_c sign\left(\boldsymbol{\dot{q}}\right)$ accounts for the motors frictions, and  $B_{eq} \left( \boldsymbol{q} \right) = B \left( \boldsymbol{q} \right) + K_r^2B_{m}$, with $B_m$ equals to the diagonal matrix of the rotor inertias; the $K_{eq}$ matrix is defined as $K_iK_r$, where $K_i$ is the diagonal matrix containing the torque-current coefficients of the motors.

The interface with the robot is based on ROS \cite{ROS}, through the \textit{ur\_modern\_driver}\footnote{https://github.com/ThomasTimm/ur\_modern\_driver}, and data are acquired with a sampling time of $8\cdot10^{-3}\,sec$. The driver provides joints positions, velocities, and currents, while accelerations are computed through causal numerical differentiation. The dataset collected is publicly available, and it has been designed to stress generalization properties. 
The training set accounts for $40000$ samples, collected through a random exploration of the robot workspace, requiring the end-effector to reach a series of random points with variable velocity. Instead, the test dataset is composed of two types of trajectories, for a total number of $25312$ points. $22324$ points have been collected through a random exploration, similar to the one described for the training dataset. The remaining samples come from the trajectory obtained requiring the end-effector to track a circle of radius $30\,cm$ at a tool speed of $30\,mm/s$.

The optimization procedures and the considered estimators are the same of the previous experiment. Due to space constraints, we neglect the FE, which performs similarly to PP. Given the higher complexity of the UR10 inverse dynamics, the number of hidden units of the neural network was increased to $600$. Concerning the derivation of the GPR based estimators, i.e. the computation of \eqref{eq:meanGP} and \eqref{eq:alphaGP}, we used a subset of the training data, in order to reduce the computational burden induced by \eqref{eq:alphaGP}; in particular, we selected $4000$ samples, downsampling with constant step the  $40000$ training samples.
The kinematic parameters considered in the derivation of the model-driven components are the nominal values provided by the manufactures.

\begin{figure}
	\centering
	\includegraphics[width=1\linewidth]{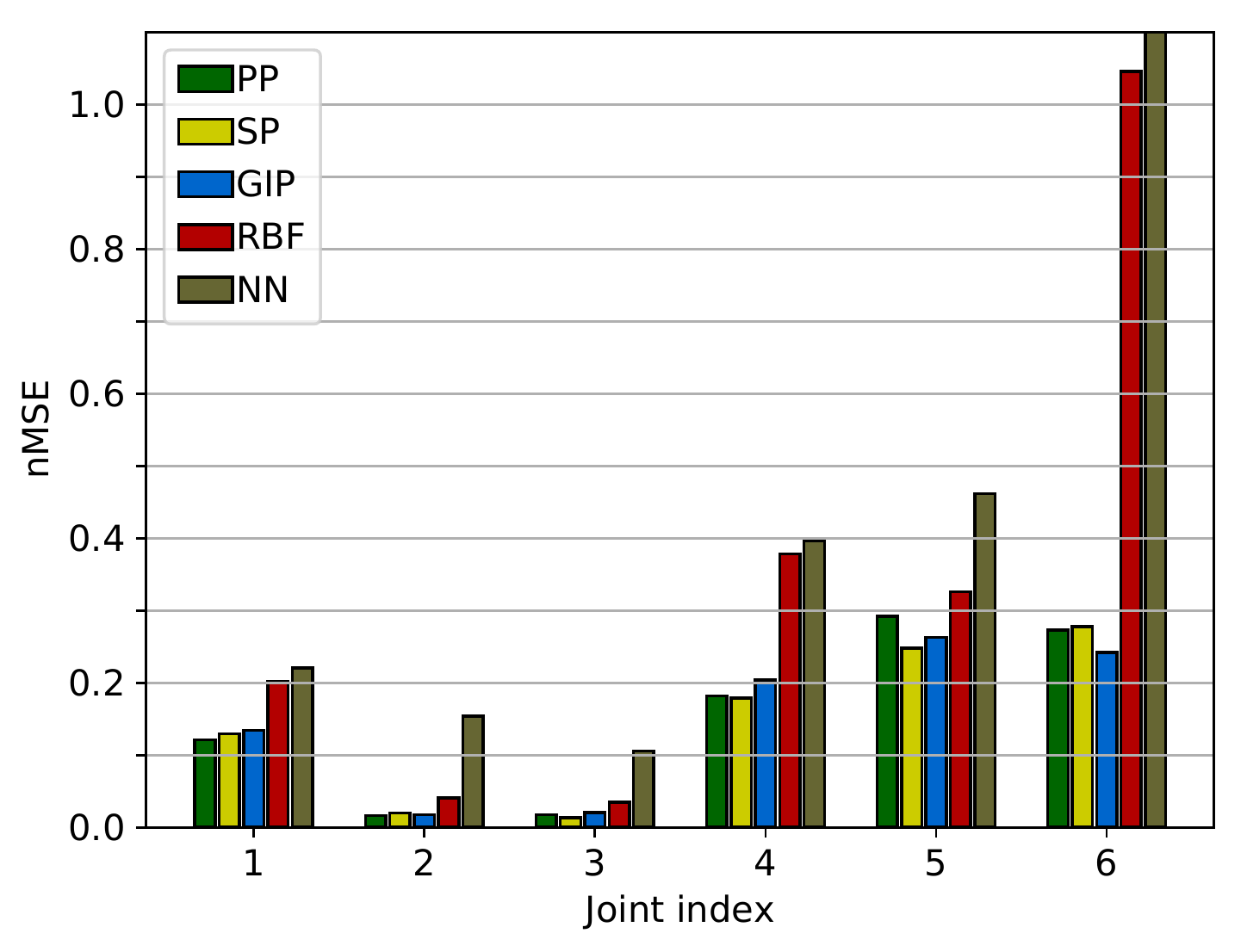}
	\caption{Comparison of the neural network (NN) and the PP, RBF and GIP kernel based estimators in the UR10 inverse dynamics prediction.}
	\label{fig:UR10_accuracy}
\end{figure}

The results obtained in the real setup, and reported in Figure \ref{fig:UR10_accuracy}, confirm the behaviors obtained in the simulative setup. The proposed approach outperforms the other data-driven estimators in all the joints, confirming that data efficiency is crucial to derive inverse dynamics estimators with good generalization properties. GIP performance is close to the ones of the model-based estimators, and in joints $5$ and $6$ the proposed approach slightly outperforms the PP estimator, that, as explained before, might be affected by model errors. SP performance confirms that model errors can be compensated by the data-driven component, even though, as proven by the $nMSE$ in joint 6, the improvement might not be so significant when data-driven estimates are not accurate.

\section{CONCLUSION} \label{sec:conclusion}
In this paper, we have introduced a novel polynomial kernel to deal with the data-driven inverse dynamics identification. Compared to other data-driven approaches, the proposed kernel, called GIP kernel, is more data-efficient. As proven by experiments in a simulated environment and also in a real system, this property allows deriving accurate inverse dynamics estimators without the need of prior knowledge about the model. Numerical results show that the GIP kernel-based estimator exhibits behaviors similar to the ones of model-based approaches, in terms of accuracy, generalization and data efficiency. However, compared to model-based solutions, our solution is not affected by model bias. Additionally, our algorithm is convenient from an implementation point of view, given its generality and hence the possibility of applying the same approach to different physical systems.


%

\appendices
\section{Proof of the Proposition \ref{prop:inv_dyn_poly}}
We prove Proposition \ref{prop:inv_dyn_poly} by inspection, analyzing individually all the terms in \eqref{eq:dyn}, i.e., the $B\left(\boldsymbol{q}\right)\boldsymbol{\ddot{q}}$ and $C\left(\boldsymbol{q},\boldsymbol{\dot{q}}\right)\boldsymbol{\dot{q}}$ contributions and the gravity term $\boldsymbol{g}\left(\boldsymbol{q}\right)$. Firstly, we provide a characterization of the elements of $B\left(\boldsymbol{q}\right)$  as polynomials in $\boldsymbol{q}_c$, $\boldsymbol{q}_s$ and $\boldsymbol{q}_p$. The inertia matrix is given by
\begin{equation*}
B\left(\boldsymbol{q}\right) = \sum_{i=1}^{i=n}m_i J_i^TJ_i + J_{\omega_i}^T R_i^0 I_i^i R_0^i J_{\omega_i} \text{,}
\end{equation*}
where $m_i$ and $I_i^i$ are the $\emph{i}$-th link mass and inertia matrix, expressed in a reference frame (RF) solidal with the $\emph{i}$-th link. $J_i$ and $J_{\omega_i}$ are the linear and angular Jacobians of the $\emph{i}$-th RF, i.e., $\boldsymbol{\dot{c}}_i = J_i \boldsymbol{\dot{q}}$ and $\boldsymbol{\omega}_i = J_{\omega_i}\boldsymbol{\dot{q}}$, where $\boldsymbol{c}_i$ is the position of the center of mass of the $\emph{i}$-th link, while $\boldsymbol{\omega}_i$ is the angular velocity of the $\emph{i}$-th RF. To expand the $J_i$ and $J_{\omega_i}$ expressions, we introduce some notions regarding the kinematics. Adopting the Denavit-Hartenberg (DH) convention, the $R_i^{i-1}$ and $\boldsymbol{l}_i^{i-1}$ variables, which denote the $\emph{i}$-th RF orientation and translation with the respect to the previous RF, are
\begin{eqnarray*}
	&&R_i^{i-1} = R_z\left(\theta_i\right)R_x\left(\alpha_i\right) \text{,}\\
	&&\boldsymbol{l}_i^{i-1} = \left[0\,,\,0\,,\,d_i\right]^T+ R_z\left(\theta_i\right) \left[a_i\,,\,0\,,\,0\right]^T\text{,}
\end{eqnarray*}
where $R_x$ and $R_z$ are the elementary rotation matrices around the $x$ and $z$ axis, while $a_i$ and $\alpha_i$ are two constant geometrical parameters, see \cite{Siciliano}. The definitions of $d_i$ and $\theta_i$ depend on the joint interconnecting the $\emph{i}$-th link with the previous link. When the joint is revolute, $d_i$ is constant and $\theta_i = \theta_{0_i} + q_i$, and the only terms that depend on $\boldsymbol{q}$ are $\cos\left(q_i\right)$ and $\sin\left(q_i\right)$ contained in $R_i^{i-1}$. Referring to the polynomial notation previously introduced, we can write that the elements of $R_i^{i-1}$ are functions in $\mathbb{P}_{(1)} \left( \cos\left(q_i\right)_{(1)},\sin\left(q_i\right)_{(1)}\right)$. In case the joint is prismatic, $\theta_i$ is constant, and $d_i = d_{0_i} + q_i$. Consequently the only $\boldsymbol{q}$ dependent terms are in $\boldsymbol{l}_i^{i-1}$. In particular the elements of $\boldsymbol{l}_i^{i-1}$ belong to $\mathbb{P}_{(1)} \left( q_{i_{(1)}} \right)$.

The $J_{\omega_i}$ matrix relates $\boldsymbol{\dot{q}}$ with the angular velocity of the \emph{i}-th link. Adopting the DH convention, $\boldsymbol{\omega}_i^{i-1} = \lambda_i\left[0\,,\,0\,,\, \dot{q}_i\right]^T$, with  $\lambda_i=1$ if the joint is revolute, and $\lambda_i=0$ if it is prismatic. Then, summing all the angular velocities projected in the base frame through the $R_j^0=\prod_{b=j}^{b=0}R_b^{b-1}$ matrices, and remarking that $\boldsymbol{\omega}_i = \sum_{j=1}^{j=i} \lambda_j R_{j-1}^0\boldsymbol{\omega}_j^{j-1}$, we obtain
\begin{eqnarray*}
	\boldsymbol{\omega}_i&=&\left[ R_0^0 \begin{bmatrix}0 \\ 0 \\ \lambda_1\end{bmatrix}
	\,,\, \dots \,,\, R_{i-1}^0\begin{bmatrix}0 \\ 0 \\ \lambda_i\end{bmatrix}\,,\,\boldsymbol{0}\left(3,n-i\right)\right] \boldsymbol{\dot{q}} \text{,}
\end{eqnarray*}
where $\boldsymbol{0}\left(3,n-i\right)$ is a $3\times\left(n-i\right)$ matrix containing only zero elements. The last equation implies 
\begin{equation*}
J_{\omega_i} = \left[ R_0^0 \begin{bmatrix}0 \\ 0 \\ \lambda_1\end{bmatrix}
\,,\, \dots \,,\, R_{i-1}^0\begin{bmatrix}0 \\ 0 \\ \lambda_i\end{bmatrix}\,,\,\boldsymbol{0}\left(3,n-i\right)\right]  \text{.}
\end{equation*}
Exploiting the properties of the rotation matrices, we obtain
\begin{equation*}
R_0^i J_{\omega_i} = \left[ R_0^i \begin{bmatrix}0 \\ 0 \\ \lambda_1\end{bmatrix}
\,,\, \dots \,,\, R_{i-1}^i\begin{bmatrix}0 \\ 0 \\ \lambda_1\end{bmatrix}\,,\,\boldsymbol{0}\left(3,n-i\right)\right] \boldsymbol{\dot{q}} \text{.}
\end{equation*}
Let $\left\{I_{r} \leq i\right\}$ be the set of revolute joint indexes lower or equal than $\emph{i}$, and let $\boldsymbol{q}_c\left( \left\{I_{r} \leq i\right\} \right)$ be the corresponding subset. Recalling that the elements of $R_i^{i-1}$ are functions in $\mathbb{P}_{(1)} \left( \cos\left(q_i\right)_{(1)},\sin\left(q_i\right)_{(1)}\right)$, with maximal degree one, and that $R_j^k = \prod_{b=j}^{b=k}R_b^{b-1}$, with $j>k$, it follows that the $J_{\omega_i}^T R_i^0 I_i^i R_0^i J_{\omega_i}$ elements belong to $\mathbb{P}_{(2 \left|\left\{I_{r} \leq i\right\}\right|)}\left(\boldsymbol{q}_c\left( \left\{I_{r} \leq i\right\} \right)_{(2)}, \boldsymbol{q}_s\left( \left\{I_{r} \leq i\right\} \right)_{(2)}\right)$, where in each monomial the following constraint holds
\begin{equation}
deg\left(q_{c_b}\right)+deg\left(q_{s_b}\right) \leq 2 \label{eq:mon_constr_2} \text{,}
\end{equation}

To derive a similar characterization of the $J_i$ elements, we analyze the $\boldsymbol{c}_i$ expression. The position of the $\emph{i-th}$ center of mass in the base frame is $\boldsymbol{c}_i = \sum_{j=1}^{j=i-1}R_{j-1}^0\boldsymbol{l}_j^{j-1} + R_i^0\boldsymbol{c}_i^i$.
Then the $\boldsymbol{c}_i$ elements are functions in $\mathbb{P}_{(i)}\left(\boldsymbol{q}_c\left( \left\{I_{r} \leq i\right\} \right)_{(1)}, \boldsymbol{q}_s\left( \left\{I_{r} \leq i\right\} \right)_{(1)}, \boldsymbol{q}_p\left( \left\{I_{p} \leq i\right\} \right)_{(1)}\right)$, and in each monomial the following inequality holds
\begin{equation}
deg\left(q_{c_{j}}\right)+deg\left(q_{s_{j}}\right) \leq 1 \label{eq:mon_constr_1} \text{.}
\end{equation} 

Since $\boldsymbol{\dot{c}}_i = J_i \boldsymbol{\dot{q}}$, and since the derivative of $\cos\left(q_j\right)$, $\sin\left(q_j\right)$ and $q_j$  does not increase the degree of these terms when inequality \eqref{eq:mon_constr_1} holds, it follows that the $J_i$ elements belong to the same functional space of $\boldsymbol{c}_i$. Consequently, the elements of $J_i^TJ_i$ are functions in  $\mathbb{P}_{(2i)}\left(\boldsymbol{q}_c\left( \left\{I_{r} \leq i\right\} \right)_{(2)}, \boldsymbol{q}_s\left( \left\{I_{r} \leq i\right\} \right)_{(2)}, \boldsymbol{q}_p\left( \left\{I_{p} \leq i\right\} \right)_{(2)} \right)$; as before, in each monomial the $q_{c_j}$ and $q_{s_j}$ degrees are subject to inequality \eqref{eq:mon_constr_2}.

Given the characterization of $J_i^TJ_i$ and $J_{\omega_i}^TR_i^0I_i^iR_0^iJ_{\omega_i}$, we obtain that the $B\left(\boldsymbol{q}\right)$ elements are functions in $\mathbb{P}_{(2n)}\left(\boldsymbol{q}_{c_{(2)}}, \boldsymbol{q}_{s_{(2)}}, \boldsymbol{q}_{p_{(2)}}\right)$, where in each monomial the $q_{c_j}$ and $q_{s_j}$ degrees are subject to inequality \eqref{eq:mon_constr_2}. Then the $B\left(\boldsymbol{q}\right)\boldsymbol{\ddot{q}}$ are functions in $\mathbb{P}_{(2n+1)}\left(\boldsymbol{q}_{c_{(2)}}, \boldsymbol{q}_{s_{(2)}}, \boldsymbol{q}_{p_{(2)}}, \boldsymbol{\ddot{q}}_{(1)}\right)$.

As reported in \cite{Siciliano}, the $\emph{i-th}$ element of the $C\left(\boldsymbol{q},\boldsymbol{\dot{q}}\right)\boldsymbol{\dot{q}}$ product is equal to
\begin{equation*}
\sum_{j=1}^{j=n}c_{ij}\dot{q}_j = \sum_{j=1}^{j=n}\sum_{k=1}^{k=n}\left(\frac{\partial b_{ij}}{\partial q_k}- \frac{1}{2}\frac{\partial b_{ik}}{\partial q_i} \right) \dot{q}_k \dot{q}_j \text{.}
\end{equation*}
Since the $B\left(\boldsymbol{q}\right)$ elements belong to $\mathbb{P}_{(2n)}\left(\boldsymbol{q}_{c_{(2)}}, \boldsymbol{q}_{s_{(2)}}, \boldsymbol{q}_{p_{(2)}}\right)$ and \eqref{eq:mon_constr_2} holds true, also the $b_{ij}$ partial derivatives belong to $\mathbb{P}_{(2n)}\left(\boldsymbol{q}_{c_{(2)}}, \boldsymbol{q}_{s_{(2)}}, \boldsymbol{q}_{p_{(2)}}\right)$, with \eqref{eq:mon_constr_2} satisfied. Indeed, for each monomial in $b_{ij}$, the derivation respect to $q_{k}$ with $k \in I_p$ decreases the degree by one, while the derivation respect to $q_k$ with $k \in I_r$ does not alter the monomial degree, being that each $q_k$ appears as $\cos(q_k)$ or $\sin(q_k)$. Then, we obtain that the elements of $C\left(\boldsymbol{q},\boldsymbol{\dot{q}}\right)\boldsymbol{\dot{q}}$ are functions in $\mathbb{P}_{(2n+1)}\left(\boldsymbol{q}_{c_{(2)}}, \boldsymbol{q}_{s_{(2)}}, \boldsymbol{q}_{p_{(2)}}, \boldsymbol{\dot{q}}_{v_{(1)}}\right)$.


Regarding $\boldsymbol{g}\left(\boldsymbol{q}\right)$, we observe that the $\emph{i-th}$ element is given by $-\partial \emph{U}/\partial q_i$, where by definition the potential energy  $\emph{U}=\sum_{j=1}^{j=n}\boldsymbol{g}_0^T\boldsymbol{c}_j$, with $\boldsymbol{g}_0$ denoting the vector of the gravitational acceleration. Then, the elements of $\boldsymbol{g}\left(\boldsymbol{q}\right)$  are functions in the same space of the $J_n$ elements.

To conclude the proof, we just need to sum all the contributions and to note that, for each joint, the torque is a function in $\mathbb{P}_{(2n+1)}\left(\boldsymbol{q}_{c_{(2)}}, \boldsymbol{q}_{s_{(2)}}, \boldsymbol{q}_{p_{(2)}}, \boldsymbol{\dot{q}}_{v_{(1)}}, \boldsymbol{\ddot{q}}_{(1)}\right)$, with each monomial satisfying \eqref{eq:mon_constr_2}.

\ifCLASSOPTIONcaptionsoff
  \newpage
\fi

\bibliographystyle{IEEEtran}
\bibliography{references}

\end{document}